%% file: acl_latex.tex
\pdfoutput=1

\documentclass[11pt]{article}

\usepackage[preprint]{acl}
\usepackage{times}
\usepackage{latexsym}

\usepackage[T1]{fontenc}

\usepackage[utf8]{inputenc}

\usepackage{microtype}

\usepackage{inconsolata}
\usepackage{pifont}
\usepackage{algorithm}
\usepackage{algorithmic}
\usepackage[namelimits]{amsmath} 
\usepackage{amssymb}             
\usepackage{amsfonts}            
\usepackage{mathrsfs}            
\usepackage{graphicx} 
\usepackage{subfigure} 
\usepackage{xcolor}

\usepackage{makecell}
\usepackage{booktabs}
\usepackage{multirow}
\usepackage{enumitem} 
\usepackage{newfloat}
\usepackage{listings}
\usepackage{subcaption}
\usepackage{xspace}
\usepackage{graphicx}
\usepackage{caption}
\usepackage{subcaption}

\usepackage{amsmath}
\usepackage{mathtools}
\newcommand{\piref}{\pi_\text{ref}}

\definecolor{green}{RGB}{36, 214, 36}
\definecolor{red}{RGB}{235, 30, 30}

\usepackage[utf8]{inputenc}
\usepackage[T1]{fontenc}

\usepackage{CJKutf8}

\usepackage{CJKutf8}
\usepackage{booktabs} 

\usepackage{CJK}
\usepackage{xcolor}
\usepackage{ulem}

\usepackage{soul}
\newcommand{\kaiti}[1]{\begin{CJK*}{UTF8}{gkai} #1 \end{CJK*}}
\soulregister{\kaiti}7
\definecolor{MyYellow}{rgb}{254, 246, 170}
\definecolor{MyBlue}{rgb}{170, 217, 251}

\title{MAPO: Advancing Multilingual Reasoning through Multilingual-Alignment-as-Preference Optimization}

\author{
    Shuaijie She \text{,} \textbf{Wei Zou} \text{,} \textbf{Shujian Huang}\thanks{Corresponding author}  \text{,} \textbf{Wenhao Zhu}\\\textbf{Xiang Liu}\text{,} \textbf{Xiang Geng}\text{,} \textbf{Jiajun Chen} \\
     \text{National Key Laboratory for Novel Software Technology, Nanjing University}  \\
    \normalsize\texttt{\{shesj,zouw\}@smail.nju.edu.cn},
    \normalsize\texttt{huangsj@nju.edu.cn}\\
    \normalsize\texttt{\{zhuwh,liuxiang,gx\}@smail.nju.edu.cn},
    \normalsize\texttt{chenjj@nju.edu.cn} \\
}

\begin{document}
\maketitle
\begin{abstract}

Though reasoning abilities are considered language-agnostic, existing LLMs exhibit inconsistent reasoning abilities across different languages, e.g., reasoning in the dominant language like English is superior to other languages due to the imbalance of multilingual training data. To enhance reasoning abilities in non-dominant languages, we propose a Multilingual-Alignment-as-Preference Optimization framework~(MAPO), aiming to align the reasoning processes in other languages with the dominant language. Specifically, we harness an off-the-shelf translation model for the consistency between answers in non-dominant and dominant languages, which we adopt as the preference for optimization, e.g., Direct Preference Optimization (DPO) or Proximal Policy Optimization (PPO). Experiments show that MAPO stably achieves significant improvements in the multilingual reasoning of various models on all three benchmarks (MSVAMP +16.2\%, MGSM +6.1\%, and MNumGLUESub +13.3\%), with improved reasoning consistency across languages~\footnote{The project will be available at \url{https://github.com/NJUNLP/MAPO}}.
\end{abstract}

\section{Introduction}
\input{latex/1-intro}

\section{Preliminary}
\input{latex/1.5-relatedwork}

\section{Method}

\input{latex/2-method}

\section{Experiment}

\input{latex/3-experiment}

\section{Experiment Results\protect\footnotemark}
\footnotetext{To facilitate further research, we launch a multilingual reasoning leaderboard: \url{https://huggingface.co/spaces/kevinpro/Open-Multilingual-Reasoning-Leaderboard}}
\input{latex/3.5-experiment-result}

\section{Analysis}
\input{latex/4-analysis}

\section{Conclusion}

In this paper, we propose MAPO, a novel multilingual alignment-as-preference optimization framework, enhancing reasoning ability in non-dominant languages by aligning them with dominant languages. 
Experimental results demonstrate that our framework achieves significant improvements on various base models across all three benchmarks, especially with a notable 16.2\% increase in average accuracy on the out-of-domain datasets MSVAMP.
The analysis confirms that enhancing alignment through our method is the key to improvements in multilingual reasoning capabilities. 

\section*{Limitation}
Similar to previous work on preference optimization, our method necessitates a policy model that has the preliminary multilingual reasoning capabilities through SFT. Meanwhile, due to limitations in computational resources, our experiments are confined to aligning models of 7B and 13B size and exploring two preference optimization algorithms, PPO and DPO. Should resources permit, we aim to extend our exploration to models of 70B sizes and examine the performance of a broader spectrum of preference optimization algorithms.


\bibliography{custom}

\appendix
\clearpage
\label{sec:appendix}
\section{Experiment Details}\label{TrainingDetails}
Our code is primarily based on the trl~\footnote{ \url{https://github.com/huggingface/trl}}
, with some minor modifications made. The modified code will also be made available at our project.

\noindent\textbf{Prompt}:
During the sampling, training, and testing phases, we consistently use the same prompt as MathOctopus~\cite{MathOctopus}.

\noindent\textbf{LoRA}:
For experiments using LoRA, such as PPO LoRA, we optimize the [q\_proj, v\_proj , o\_proj]  modules. The values of r and alpha are set to 128 and 256, respectively.

\noindent\textbf{DPO}: 
We employ a learning rate of 1e-6, with $\beta$ set at 0.1, and a warmup step count of 100. The batch size is configured to 128. The optimization process is capped at a maximum of 1000 steps, and we save the checkpoint that corresponds to the lowest loss on validation set.
The training takes around 4 hours on 8 H100 GPUs.

\noindent\textbf{PPO}: 
We have configured a learning rate of 2e-5, with a batch size of 64 and 2 ppo epochs. We utilize the AdamW optimizer to improve the stability of the optimization, setting the epsilon value to 1e-5. Additionally, we have implemented a linear learning rate warm-up technique for the first 150 steps. All other hyper parameters are set to the default values provided by the trl library. We set the maximum optimization steps to 2600 and report the results at this checkpoint.

\section{Supplemental Experiment Results}\label{SupplementResult}
To verify the robustness of our method, we conducted experiments on different preference optimization algorithms. Due to the limited computational resources, we optimized the PPO algorithm using LoRA. The detailed experimental results of PPO LoRA and each round in Iterative DPO are shown in Table~\ref{Tab:ALLPRMethodResult}. 

Experiments demonstrate that our framework achieve an effective improvement in multilingual reasoning capabilities based on both PPO and DPO. Despite the limited computational resource, where PPO only updated a part of parameters with merely 2600 steps, it has already brought impressive performance enhancements on all three datasets. Additionally, with the increasing rounds of DPO, the model exhibited progressively stronger multilingual reasoning abilities that has not yet reached its limit, revealing the potential of our approach.

\section{Used Scientific Artifacts}
Below lists scientific artifacts that are used in our work. For the sake of ethic, our use of these artifacts is consistent with their intended use.
\begin{itemize} [itemsep=1pt]
    \item \textit{Transformers (Apache-2.0 license)}, a framework to facilitate downloading and training state-of-the-art pretrained models.
    \item \textit{trl (Apache-2.0 license)}, a full stack library where provide a set of tools to train transformer language models with Reinforcement Learning. The library is built on top of the Transformers library.
\end{itemize}

\begin{figure}[t]
\centering
\includegraphics[scale=0.26]{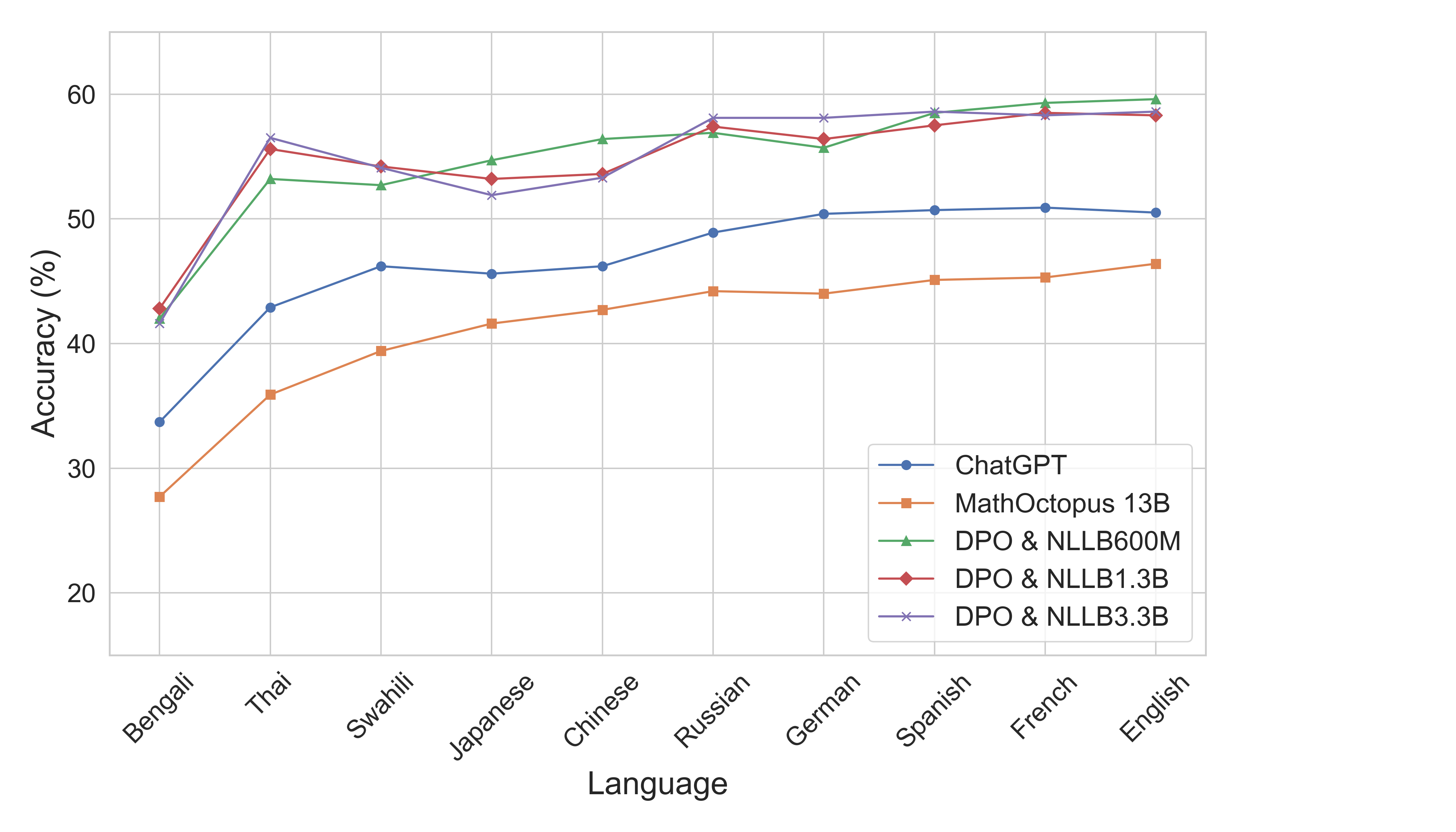}
\caption{Accuracy across ten languages on MSVAMP after training MathOctopus 7B on preference datasets constructed using translation models of different scales.}
\label{ScaleAblation}
\end{figure}

\section{Scaling of Translation Model}\label{ScalingMT}
We also conducted ablation experiments comparing the translation models used during the preference estimation phase. In addition to NLLB-600M-distilled, we employed NLLB-1.3B-distilled 
 and NLLB-3.3B 
 to estimate the preference and repeat the experiments. The results are shown in Figure~\ref{ScaleAblation}, where we find that all three models achieved impressive results, showing stable improvements over the Base Model and ChatGPT across ten languages. These results demonstrate that our method is robust for translation models of various sizes.

\begin{table*}[t]
\setlength{\tabcolsep}{3mm}
\footnotesize
\centering
\begin{tabular}{l|cccccccccc | c}\toprule
\textbf{Model} & \textbf{Bn} & \textbf{Th}	& \textbf{Sw}	& \textbf{Ja} & \textbf{Zh}	& \textbf{Ru}	& \textbf{De}	& \textbf{Es}	& \textbf{Fr}	   & \textbf{En} & \textbf{Avg}  \\\midrule 
GPT-3.5-Turbo   & 36.2 & 42.6 & 47.2 & 58.1 & 60.6 & 42.6 & 41.5 & 54.9 & 39.4 & 70.6 & 49.4\\
\midrule
MAmmoTH 7B & 1.7&	4.7	&4.5	&23.5	&29.9	&26.9	&34.7	&37.9	&36.5	&42	&24.2 \\
WizardMath 7B & 9.8	 & 13	& 6.2	& 31.8&	36	&34.3	&34.1	&39.9	&36.5	&45.4	&28.7	\\
MetaMath 7B & 13.2	&19.2	&12.1&	50.7	&53.3	&52.9&	54.2&	56.5	&56.7	&63.3	&43.2\\
\midrule
MathOctopus 7B & 26.6 & 30.9 & 34.3 & 40.9 & 44.4 & 36.0 & 32.6 & 42.0 & 36.2 &  46.9 & 37.1\\
 + m-RFT & 38.0 & 42.9 & 41.8 & 48.2 & 51.6 & 45.2 & 42.9 & 49.3 & 42.9 & 51.2 & 45.4\\
+ MAPO-DPO(ours) & 41.8 & 45.8 & 46.9 & 52.9 & 54.4 & 49.9 & 50.7 & 54.0 & 51.4 & 55.9 & 50.4 \\\midrule
MetaMathOctopus 7B& 24.5 & 35.2 & 36.2 & 44.1 & 45.8 & 39.2 & 32.6 & 45.2 & 36.7 &52.5 & 39.2\\
 + m-RFT & 40.3	&45.0	&45.2&	51.4	&57.8	&51.6	&51.6&	58.8	&50.1	& 65.3 & 51.7	 \\
+ MAPO-DPO(ours) & 40.7	& 46.0	& 45.0	& 58.2	& 59.3	& 53.1	& 51.4	& 57.4	& 52.0 & 	66.1 & 52.9\\\midrule 
MistralMathOctopus 7B & 45.2&	50.7	&49.5	&56.5	&65.3	&59.1&	51.4&	62.1	&53.9	&74.6&	56.8	 \\
+ m-RFT & 52.7&	60.6	&61.0	&67.2	&70.8&	65.9	&61.2	&71.6&	64.4&	78.3	&65.4	 \\ 
+ MAPO-DPO(ours) & \textbf{62.3}	& \textbf{64.6}	& \textbf{61.6}	& \textbf{72.1}	& \textbf{75.1}	& \textbf{68.0}&	\textbf{69.3}&	\textbf{74.4}	&\textbf{74.2}	&\textbf{78.5} &	\textbf{70.0}	\\
\midrule \midrule 

MAmmoTH 13B & 6.8&	10.5	&10.4	&31.6&	38	&41.1&	41.4	&43.3&	42.6&	46.7	&31.2	 \\
WizardMath 13B & 9.8&	14.3	&12.4	&30.5&	39.0	&36.5	&35.2	&43.9&	39.2	&47.8&	30.9 \\
MetaMath 13B & 10.9&	16.0	&16.0	&55.2	&57.4	&56.5&	58.9&	60.6	&58.0	&64.2	&45.4 \\
\midrule
MathOctopus 13B &42.4&	39.2&	44.8	&38.8	&49.6	&45.2&	48.4&	53.6	&43.2		&54.8 & 46.0 \\
+ m-RFT & 44.1	&49.9&	51.0	&51.0&	55.6	&49.0	&47.1	&53.3	&46.1&	57.4	&50.5	 \\ 
+ MAPO-DPO(ours)  &49.0	&53.3	&52.2&	55.0	&57.4&	53.3	&52.2&	55.9	&50.7&	58.8&	53.8	\\
\midrule
MetaMathOctopus 13B &  34.4&	42.8&	41.6&	49.2	&52.8	&54.4&	54.4	&59.2&	53.6	&71.6&	51.4\\
+ m-RFT &40.1&	51.6	&47.5	&60.5	&62.0&	58.9	&54.8	&62.5	&54.2	&66.5	&55.9		\\
+ MAPO-DPO(ours) & \textbf{52.9}	& \textbf{55.4}&	\textbf{55.0}&	\textbf{67.2}&	\textbf{65.0}&	\textbf{54.8}&	\textbf{54.4}&	\textbf{65.2}&	\textbf{57.4}	&\textbf{70.4}&	\textbf{59.8}\\
\bottomrule
\end{tabular}
\caption{\label{Tab:AllLanResult-NumGLUE} Model Performances on MNumGLUESub test set. ``Avg'' represents the average performance in ten languages and bold text denotes the best results within the same model size.}
\end{table*}

\begin{table*}[t]
\setlength{\tabcolsep}{3mm}
\footnotesize
\centering
\begin{tabular}{l|cccccccccc | c}\toprule
\textbf{Model} & \textbf{Bn} & \textbf{Th}	& \textbf{Sw}	& \textbf{Ja} & \textbf{Zh}	& \textbf{Ru}	& \textbf{De}	& \textbf{Es}	& \textbf{Fr}	    & \textbf{En}   & \textbf{Avg}\\\midrule 
& \multicolumn{11}{c}{\textit{Benchmark: MSVAMP}} \\\midrule
MathOctopus 7B  &27.7&	35.9&	39.4&	41.6	&42.7	&44.2&	44.0	&45.1&	45.3	&46.4	&41.2	\\
+ PPO LoRA & 38.9&	47.5&	47.1&	51.0	&51.7&	51.1&	50.3&	51.6&	51.4&	52.7&	49.3\\
+ DPO Iter1& 42.0	&53.2	&52.7	&54.7	&56.4	&56.9&	55.7	&58.5	&59.3&	59.6	&54.9\\
+ DPO Iter2 & 45.3	&53.9	&53.8	&56.8	&58.1&	56.6	&58.7	&59.1	&58.5&	60.0	&56.1\\
+ DPO Iter3 & 48.8	&55.2&	56.0	&60.3&	58.8&	58.3	&58.1&	59.7&	60.8	&58.4&	57.4	\\\midrule
MetaMathOctopus 7B & 36.1& 	47.5	& 49.4& 	51.3& 	54.5	& 53.6& 	56.6	& 60.0	& 57.2	& 64.2	& 53.0	\\
+ PPO LoRA & 45.7	&52.0	&52.2	&61.2	&58.3	&57.0&	58.7	&60.7&	61.7	&67.5	&57.5	\\
+ DPO Iter1 & 44.1&	58.2&	59.0	&60.3&	62.3&	63.5	&65.1	&63.7&	64.2&	70.2&	61.1\\
+ DPO Iter2 & 48.5	&61.8	&59.2&	64.3	&64.3&	64.4	&65.0 &	66.1 &	65.4&	70.9&	63.0\\
+ DPO Iter3 & 50.1	&61.6&	61.7&	65.9	&65.7&	64.8&	68.4	&68.5	&68.6&	71.6	&64.7\\\midrule
& \multicolumn{11}{c}{\textit{Benchmark: MGSM}} \\\midrule
MathOctopus 7B  &29.2&	33.6&	36.4&	35.2&	39.2&	38.8	&44.8&	42.4	&43.2	&52.0	&39.5	\\
+ PPO LoRA & 31.2&	38.4&	38.4	&37.2&	43.6&	35.2	&46.0	&44.0	&38.8&	51.2&	40.4\\
+ DPO Iter1& 29.2 & 	36.4& 	35.6& 	35.6	& 41.6	& 38.4& 	40.8	& 42.0& 	37.6& 	46.8& 	38.4\\
+ DPO Iter2 & 30.4	&36.0	&37.6&	38.0	&45.2	&39.6&	42.0	&47.6	&41.2&	45.2&	40.3\\
+ DPO Iter3 & 30.8	&38.0	 &37.6	&45.2&	47.2	&42.0	&45.2&	43.2	&40.8&	45.6&	41.6\\\midrule
MetaMathOctopus 7B & 25.6&	42.8&	36.4&	40.0&	46.4&	46.8&	49.6	&54.4&	46.4	&66.4&	45.5	\\
+ PPO LoRA & 36.0 &	41.2&	41.6	&46.4	&54.8&	53.6&	54.0&	55.6	&51.6&	68.0&	50.3	\\
+ DPO Iter1 & 32.8	&43.2&	40.4&	48.8	&49.2&	52.8&	54.4	&52.8&	50.0&	64.8&	48.9\\
+ DPO Iter2 & 34.0&	48.0&	45.2&	40.4&	54.0&	52.0&	50.8&	54.0&	49.2&	70.4&	49.8\\
+ DPO Iter3 & 36.0 &	44.8	&44.8	&47.6&	55.2	&53.6	&53.6&	56.8	&52.4	&70.8	&51.6\\\midrule
& \multicolumn{11}{c}{\textit{Benchmark: MNumGLUESub}} \\\midrule
MathOctopus 7B  &26.6	 &30.9	 &34.3	 &40.9	 &44.4 &	36.0 &	32.6 &	42.0 &	36.2 &	46.9 &	37.1	\\
+ PPO LoRA & 34.3&	41.1&	40.7&	45.6&	49.3	&39.5&	34.3&	46.7&	36.3&	51.6&	41.9\\
+ DPO Iter1& 32.6&	41.4	&42.0&	42.7&	45.0&	42.2&	40.7	&47.8&	40.9&	47.8&	42.3\\
+ DPO Iter2 & 35.6	&39.7	&42.2&	45.6	&50.7&	44.1&	42.6	&49.3&	42.9	&49.7&	44.2\\
+ DPO Iter3 & 41.8&	45.8	&46.9	&52.9	&54.4	&49.9&	50.7&	54.0	&51.4	&55.9&	50.4\\\midrule
MetaMathOctopus 7B & 34.7	&41.4&	37.9	&47.8&	54.2	&45	&43.1	&52.5&	45.0&	60.8&	46.3	\\
+ PPO LoRA & 42.4	&46.7	&45.4&	55.2&	58.9	&45.0&	41.6&	55.7&	44.6&	64.6&	50.0		\\
+ DPO Iter1 & 37.7&	46.3	&43.5&	54.8	&58.4&	50.5&	52.2	&60.1	&51.6	&62.1	&51.7	\\
+ DPO Iter2 & 40.7	& 46.9	&45.0	&53.5&	58.4&	50.8	&51.4	&58.4&	49.9	&64.6	&52.0	\\
+ DPO Iter3 & 40.7	&46.0&	45.0&	58.2&	59.3&	53.1&	51.4&	57.4&	52.0&	66.1&	52.9\\
\bottomrule
\end{tabular}
\caption{\label{Tab:ALLPRMethodResult} Model Performances on three benchmarks. We report the results of PPO LoRA and each round in Iterative DPO. ``Avg'' represents the average performance on ten languages}
\end{table*}
\end{document}

%% file: latex/1-intro.tex

The reasoning ability of large-scale models has attracted much attention~\cite{gsm8k,self,cot,metamath}.
Though we consider reasoning to be language-agnostic, existing studies~\cite{MathOctopus} show that due to the imbalance of pre-training and fine-tuning data across languages, the mathematical reasoning ability of existing large-scale models varies across languages, e.g., English, is far superior to that of the other languages.

To improve the reasoning ability in other languages, ~\citet{MathOctopus} translated the English reasoning processes into other languages for supervised fine-tuning~\cite[SFT]{instructgpt}. 
Although SFT brings in preliminary capability for multilingual reasoning, we argue that two problems hinder further improvement.


Firstly, the annotation for the reasoning process is expensive to obtain even for the dominant language, and the reasoning processes involve complex mathematical reasoning, which may result in translation errors~\cite{MathOctopus}. 
As a result, the translated reasoning annotation for non-dominant languages is limited in both scale and quality.
Without sufficient and diverse data, the results of supervised training are limited. 
It may also suffer from generalization issues in versatile task scenarios~\cite{secretsRLHF,instructgpt}, indicating potential difficulty in adapting to the out-of-domain test set.

More importantly, although supervised training with translated reasoning processes improves the reasoning ability for almost all trained languages, this strategy only fills in the missing reasoning annotation for non-dominant languages that originate from the dominant language. 
Thus the inherent gap between dominant and non-dominant languages is hardly narrowed.

Different from existing attempts, we propose to use the reasoning ability of the dominant language as the director for improving non-dominant languages. 
As the reasoning process is critical for obtaining the correct result, the multilingual reasoning ability may be improved by encouraging the reasoning in non-dominant languages to be similar to that in the dominant language.
Therefore, we propose a \textbf{M}ultilingual-\textbf{A}lignment-as-\textbf{P}reference \textbf{O}ptimization~(MAPO) framework by aligning the reasoning process of non-dominant languages to the dominant. 
Notably, MAPO exploits the strong reasoning ability in the dominant language and requires no annotation for the reasoning process.



More specifically, MAPO consists of two stages: preference estimation via multilingual alignment and preference optimization. 
During preference estimation, the reasoning processes to the same question are sampled from the LLM in both dominant and non-dominant languages.
A well-trained, off-the-shelf translation model is employed to yield the translation probability between the reasoning in dominant and non-dominant languages.
Since higher translation probability indicates a more consistent reasoning aligned with the dominant language, the corresponding reasoning in the non-dominant language is considered better and shall be promoted. 
During preference optimization, Proximal Policy Optimization~\cite[PPO]{ppo} and Direct Preference Optimization~\cite[DPO]{dpo} are adopted to optimize the previously estimated preference. 
So the LLMs are trained to reason in the non-dominant languages as they do in the dominant language.  
We also conduct Iterative DPO for further preference optimization.


Experiments are conducted on three challenging multilingual reasoning test sets, namely MSVAMP~\cite{MathOctopus}, MGSM~\cite{mgsm}, and MNumGLUESub constructed from NumGLUE ~\cite{numglue}, each covering 10 languages.
MAPO achieves accuracy improvements of up to 16.2\%, 6.1\%, and 13.3\% on the three benchmarks, respectively, reaching state-of-the-art performance on 7B models, thereby demonstrating the effectiveness of our method. 
Among them, the improvement on the out-of-domain dataset MSVAMP (+16.2\%) shows that, by enhancing multilingual reasoning consistency, MAPO improves the multilingual reasoning capabilities of the model in a generalizable manner.


%

%% file: latex/1.5-relatedwork.tex


\subsection{Multilingual Reasoning}
A straightforward measurement to evaluate the efficacy of large language models~(LLMs) lies in their proficiency in tackling complex reasoning, e.g., their performance in solving mathematical reasoning.
Recent work~\cite{cot,self} has verified a substantial improvement when LLMs are guided through a step-by-step reasoning process, instead of conducting a direct answer.

Some research has introduced mathematical reasoning datasets in the form of application problems, such as GSM8K~\cite{gsm8k}, NumGLUE~\cite{numglue}, and SVAMP~\cite{svamp}. To evaluate the multilingual reasoning capabilities of LLMs, \citealp{mgsm} propose MGSM by manually translating 250 samples of the GSM8K test set from English to other languages.
Subsequent researches focus on enhancing the multilingual reasoning of LLMs. 
\citealp{MathOctopus} translate the GSM8K training data into other languages for supervised fine-tuning~(SFT), which improves the model's multilingual reasoning capabilities. 
However, SFT suffers from data scarcity and catastrophic forgetting.
Its out-of-domain generalization is also hard to guarantee~\cite{secretsRLHF}.
Multilingual reasoning via LLMs remains an open challenge.



\subsection{Preference Optimization}

SFT maximizes the likelihood of annotated outputs and equips models with preliminary capabilities. However, models still exhibit various issues after SFT. Some researchers have further adjusted model behaviors and enhanced model capabilities through preference optimization.

RLHF~\cite{secretsRLHF,HHassiRLHF} further rectifies these model behaviors via reinforcement learning by preference.
RLHF introduces a reward model $r_\theta(x,y)$ given input $x$ with its corresponding output $y$ that captures the preference nuance from the human feedback.
Then, $r_\theta(x,y)$ scores arbitrary LLM outputs $y$ given input $x$ for iterative policy rectifications during proximal policy optimization~\cite[PPO]{ppo}.
The tuning is guided by Eq~\ref{eq:loss-ppo} to maximize the expected rewards of the LLM policy $\pi_\theta$ with the minimum deviation from the SFT policy:
\begin{equation}
    \mathcal{L}_{\text{PPO}}=\mathbb{E}_{(x,y)\sim D_\pi}[r_\theta(x,y) - \gamma\log\frac{\pi_\theta(y|x)}{\pi_\text{SFT}(y|x)}],
    \label{eq:loss-ppo}
\end{equation}
where $\pi_\text{SFT}$ is the original LLM via SFT, $\gamma$ is a hyperparameter that constrains policy updates.

Though RLHF via PPO is effective in adapting LLMs to versatile human preferences, it involves four sub-models, making the training complex and costly.
DPO~\cite{dpo} proposes to distill a referential SFT policy $\pi_\text{ref}$ by polarizing the preference.
DPO tuning involves a pair of outputs $(y_w, y_l)$ in Eq~\ref{eq:loss-dpo}:
\begin{flalign}
    &\mathcal{L}_\text{DPO}(\pi_{\theta}; \piref) =  -\mathbb{E}_{(x, y_w, y_l)\sim \mathcal{D}} & \nonumber \\
&\left[\log \sigma \left(\beta \log \frac{\pi_{\theta}(y_w\mid x)}{\piref(y_w\mid x)} - \beta \log \frac{\pi_{\theta}(y_l\mid x)}{\piref(y_l\mid x)}\right)\right], 
\label{eq:loss-dpo}
\end{flalign}
where $y_w$ is favored over $y_l$, and $\beta$ is a hyperparameter.

\begin{figure*}[ht]
\centering
\includegraphics[scale=0.61]{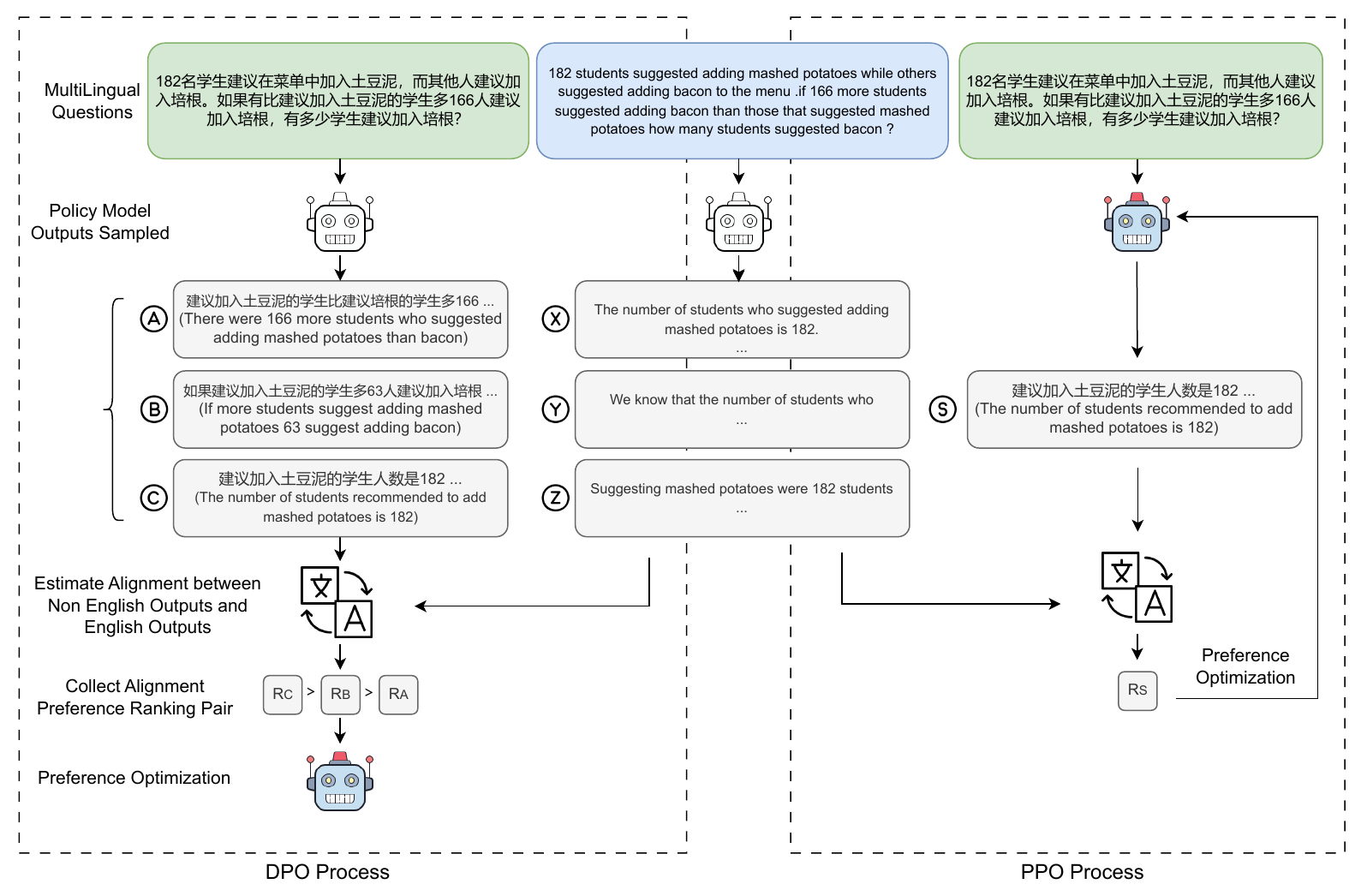}
\caption{Illustration of our alignment framework. For brevity, we only show three sampling results and simplified optimization processes of DPO and PPO. The green and blue colors represent the same problem in Chinese and English, respectively. The white robots represent the original policy model and the colored robots represent the policy model with parameters updated through preference optimization.}
\label{align_illus}
\end{figure*}

%% file: latex/2-method.tex


\subsection{Preference Estimation}
Intuitively, reasoning ability is language-agnostic.
However, LLM reasoning varies across different languages, where we consider a dominant language provides the better reference in reasoning for lesser languages.
Therefore, a straightforward approach is to align reasoning in non-dominant languages to a dominant language.
That is, we can rectify LLM's output reasoning in non-dominant languages by favoring those answers more aligned to the dominant language.

We simply adopt a well-trained multilingual translation model to calculate the `alignment' scores between answers in non-dominant and dominant languages as a preference.
Since the translation model is optimized on a large scale parallel data in source and target languages $(Y, \bar{Y})$ by maximizing conditional generation probability in Eq~\ref{eq: translation target}:
\begin{equation}
    \mathop{\arg\max}\limits_{\theta} P(\bar{Y} | {Y}; \theta), \label{eq: translation target}
\end{equation}
with more aligned target language $\bar{Y}$ to $Y$, the higher the conditional generation probability $P(\bar{Y}|Y)$.
Then, we input the answers in non-dominant languages and force-decode a corresponding answer $\bar{Y}$ in dominant language for the `alignment' score.
Correspondingly, answers with higher `alignment' scores in non-dominant languages are preferred during the following preference optimizations.




\subsection{Preference Optimization}
With `alignment' scores as a preference, we adopt two state-of-the-art preference optimizations: PPO and DPO, to optimize the alignment scores of reasoning in non-dominant languages given that in a dominant language.
To make life easier, since English is dominant in both reasoning and translation, we adopt English as the dominant language in the rest of the paper.

\subsubsection{Optimization with PPO}
In our setup, the `alignment' score $P(\bar{Y}|Y)$ given query $x$ depicts the preference as the reward model $r_\theta(x,Y)$ do in PPO, thus we directly adopt the multilingual machine translation model as the reward model.
During each optimization, we collect a batch of non-English outputs $\{Y_i\}$ with an English output $\bar{Y}$, respectively.
PPO then maximizes the expected alignment score within the batch by Eq~\ref{eq:our-ppo}:
\begin{equation}
    \mathcal{L}_{\text{PPO}}=\mathbb{E}_{i}[P(\bar{Y}|Y_i, x) - \gamma\log\frac{\pi_\theta(Y_i|x)}{\pi_\text{SFT}(Y_i|x)}].
    \label{eq:our-ppo}
\end{equation}


\subsubsection{Optimization with DPO}
DPO involves an input pair $(Y_w, Y_l)$, where $Y_w$ is favored over $Y_l$.
Correspondingly, we collect $n$ outputs within one non-English language, making $\binom{n}{2}$ pairs of $(Y_w, Y_l)$ ranked by the alignment scores given the same English answer $\bar{Y}$.
We additionally copy and freeze the original supervised fine-tuned LLM as the reference policy $\pi_\text{ref}$ in Eq~\ref{eq:loss-dpo}.
Finally, our DPO is optimized by Eq~\ref{eq:our-dpo}:
\begin{flalign}
    &\mathcal{L}_\text{DPO}(\pi_{\theta}; \piref) =  -\mathbb{E}_{(x, \bar{Y}, Y_w, Y_l)\sim \mathcal{D}} & \nonumber \\
&\left[\log \sigma \left(\beta \log \frac{\pi_{\theta}(Y_w\mid x)}{\piref(Y_w\mid x)} - \beta \log \frac{\pi_{\theta}(Y_l\mid x)}{\piref(Y_l\mid x)}\right)\right].
\label{eq:our-dpo}
\end{flalign}

\subsubsection{Iterative DPO}

%
Recent research such as LLaMA2~\cite{llama2} and Claude~\cite{anthropic} suggests that preference data updated through multiple iterative rounds can is beneficial for the performance of preference optimization.

Therefore, we further conduct Iterative DPO to optimize reasoning alignments across languages. 
Specifically, assume that initially when we train $\pi_{\theta_i}$ on preference data sampled by itself, we obtain $\pi_{\theta_{(i+1)}}$. Subsequently, we continue to sample based on $\pi_{\theta_{(i+1)}}$ and update it further on the newly sampled preference data.

%% file: latex/3-experiment.tex
\subsection{Experiment Setup\protect\footnotemark}
\footnotetext{More training details can refer to Appendix~\ref{TrainingDetails}}
\noindent\textbf{Base Models:} Our experiments include four base models. \textbf{MathOctopus-7/13B} is fine-tuned from LLaMA2-7/13B with the multilingual reasoning processes MGSM8KInstruct\footnote{GSM8K translated to nine non-English languages.}
\cite{MathOctopus}. \citealp{metamath} proposed MetaMath-7/13B and MetaMath-Mistral-7B, which have stronger English reasoning abilities. We fine-tune these models with MGSM8KInstruct and get \textbf{MetaMathOctopus-7/13B} and \textbf{MistralMathOctopus-7B}.



\noindent\textbf{Preference Estimation}:
To obtain the mathematical problems for preference optimization, we selected tasks 1, 4, and 8 from the eight tasks in NumGLUE~\cite{numglue}, which is a multi-task arithmetic reasoning benchmark, and translate the questions into the same nine languages as in MGSM8KInstruct, resulting the dataset MNumGLUESub. MNumGLUESub with only 1700 problems.
When constructing the preference dataset for DPO, we use the corresponding base model for sampling and calculate the alignment feedback using NLLB-600M-distilled~\footnote{\url{ https://huggingface.co/facebook/nllb-200-distilled-600M}}.

\noindent\textbf{Preference Optimization }:
We experimented with both PPO and DPO. For simplicity, we report the results of the third iteration of DPO by default. For detailed results of PPO-LoRA and each round in Iterative DPO, please refer to Appendix~\ref{SupplementResult}.

\begin{table}[t]
\centering
\footnotesize
\begin{tabular}{lccc}
\toprule
Dataset                & Type       & Number  \\
\midrule
MGSM    & In-Domain        & 2,500 \\
MNumGLUESub & In-Domain        & 5,530  \\
MSVAMP          & Out-of-Domain      & 10,000   \\
\bottomrule
\end{tabular}
\caption{Statistics of three benchmarks.  ``Type" indicates whether its corresponding training set was used during the training.  `` Number'' refers to the total number of samples in the test data.}
\label{tab:statistics}
\end{table}

\noindent\textbf{Evaluation Datasets}:
We utilize three challenging benchmarks: MSVAMP, MGSM~\cite{mgsm} and MNumGLUESub (statistics are shown in Table~\ref{tab:statistics}). 
Among them, MSVAMP, serving as an Out-of-domain test set that does not participate in the training, is used to analyze the model's robustness and generalization ability. 
MGSM is the testset corresponds to MGSM8KInstruct, on which the base models are trained.
MNumGLUESub is the testset corresponds to the data for preference estimation and optimization.  


\begin{table*}[t]
\setlength{\tabcolsep}{3mm}
\footnotesize
\centering
\begin{tabular}{l|cccccccccc | c}\toprule
\textbf{Model} & \textbf{Bn} & \textbf{Th}	& \textbf{Sw}	& \textbf{Ja} & \textbf{Zh}	& \textbf{Ru}	& \textbf{De}	& \textbf{Es}	& \textbf{Fr}	 & \textbf{En} & \textbf{Avg} \\\midrule 
GPT-3.5-Turbo & 33.7 & 42.9 & 46.2 & 45.6 & 46.2 & 48.9 & 50.4 & 50.7 & 50.9 & 50.5 & 46.6  \\\midrule
%

MAmmoTH 7B$^\dagger$ & 4.3  & 6.3  & 4.2  & 26.7 & 26.8 & 33.7& 39.6 &42.9 &39.9 & 45.1 & 26.3 \\
WizardMath 7B$^\dagger$ & 16.1 & 17.0 & 10.3 & 37.9 & 36.3 & 37.4 &39.2 &44.8 &37.7 & 48.5 & 32.5 \\
MetaMath 7B & 14.8 & 	17.7& 	14.8	& 51.8	& 54.4	& 59.4	& 59.6	& 63.2	& 62.0	& 64.8	& 46.2 \\
\midrule
MathOctopus 7B   &  27.7 & 35.9 & 39.4 & 41.6 & 42.7 & 44.2 & 44.0 & 45.1 & 45.3 &  46.4 & 41.2\\
+ m-RFT  &37.9	&46.4	&46.4	&49.6&	50.8&	50.4&	50.7&	51.6	&53.4	&49.4	&48.7\\
+ MAPO-DPO(ours)  & 48.8 & 55.2 & 56.0 & 60.3 & 58.8 & 58.3 & 58.1 & 59.7 & 60.8 & 58.4 & 57.4\\
\midrule 
MetaMathOctopus 7B &  36.1	 & 47.5	 & 49.4	 & 51.3	 & 54.5	 & 53.6	 & 56.6	 & 60.0	 & 57.2 & 64.2 & 53.0 \\
+ m-RFT & 44.8&	54.2&	56.2	&58.4&	57.7	&56.2&	59.2	&59.3	&57.8	&63.1	&56.7 \\ 
+ MAPO-DPO(ours) & 50.1	& 61.6	& 61.7	& 65.9	& 65.7	& 64.8	& 68.4 &	68.5	& 68.6	&  71.6 & 64.7\\\midrule 
MistralMathOctopus 7B & 47.0	&52.6	&54.4	&58.8	&60.2	&59.8	&62.1	&60.8	&60.4	&74.4&	59.0 \\
+ m-RFT & 57.4 &	63.5	&60.8	&65.2	&69.8	&67.4&	67.2	&69.1&	68.0	&71.3&	66.0 \\ 
+ MAPO-DPO(ours) & \textbf{62.9}	& \textbf{71.3} &	\textbf{71.4}	& \textbf{73.7}	& \textbf{76.0}	& \textbf{74.9}&	\textbf{77.8}	& \textbf{78.1}	&\textbf{79.0}	&\textbf{81.1}	& \textbf{74.6}\\
\midrule \midrule 
MAmmoTH 13B$^\dagger$  &5.0 &13.7 &12.9 &42.2 &47.7& 50.7 &52.3 & 53.9&  53.8& 53.4 & 38.6 \\
WizardMath 13B$^\dagger$ &  13.7& 16.3 &12.5 &29.5 &37.0& 43.8& 48.7& 49.4 &49.4 & 56.3 &35.7 \\
MetaMath 13B & 14.6 & 	15.7& 	17.4	& 57.0	& 56.6	& 63.7	& 67.3	& 65.9	& 64.7	& 67.7	& 49.1 \\
\midrule
MathOctopus 13B   & 35.2  &  46.8   &  42.8  &  43.2  &  48.8  &  47.6  &  44.4  &  48.0  &  48.4  &  53.2 & 45.8 \\
+ m-RFT  & 43.4	&50	&52.1	&54.9&	55.4	&57.1&	59.2	&56.4&	59.5&	55.2&	54.3 \\
+ MAPO-DPO(ours)  & 51.8&	58.9	&56.4&	60.4&	58.8&	62.1&	63.5	&62.0 & 61.7 &	65.0	& 60.1\\
\midrule
MetaMathOctopus 13B &  41.6&	52.1	&50.9&	57.3	&53.1&	59.1&	60.1&	61.1&	60.8	&66.8&	56.3 \\
+ m-RFT  &48.1	&59.6	&61.4&	60.5	&58.9&	61.0	&62.7	&65.3	&64.3	&65.4	&60.7\\
+ MAPO-DPO(ours) & \textbf{54.7}	& \textbf{64.7} &	\textbf{62.9}	& \textbf{69.0}	&\textbf{68.2}	& \textbf{68.2}&	\textbf{69.5}&	\textbf{70.6}&	\textbf{71.3}&	\textbf{70.5}&	\textbf{67.0}\\
\bottomrule
\end{tabular}
\caption{\label{Tab:AllLanResult-MSVAMP} Model Performances on MSVAMP test set. ``Avg'' represents the average performance on ten languages and bold text denotes the best results within the same size. Results marked with $^\dagger$ come from \citealp{MathOctopus}.}
\end{table*}

\noindent\textbf{Evaluation Metric}:
\begin{itemize}[topsep=0pt, itemsep=1pt, parsep=0pt, partopsep=0pt]
    \item \textbf{Accuracy}: We use the accuracy of problem-solving to measure the model's reasoning ability, with a higher accuracy representing stronger reasoning ability.
    \item \textbf{PPL (PPL-based Alignment Score)}: We input the target non-English answer and apply NLLB-600M-distilled to get the perplexity of the given English answer. Less perplexity indicates better alignment between the reasoning processes.
    \item \textbf{ACR (Answer Consistency Ratio) }: Let $m$ denote the set of math questions answered correctly in English, and $n$ denote those answered correctly in non-English. ACR is then calculated as: $\text{ACR} = \frac{|m \cap n|}{|n|}$. Higher ACR indicates a greater degree of overlap in reasoning capabilities between non-English and English languages.
\end{itemize}

\noindent\textbf{Baselines}:
The selected base models are already strong baselines, which have been fine-tuned on \textsc{MGSM8KInstruct}.
Moreover, motivated by Rejection sampling Fine-Tuning (RFT)~\cite{rft}, we employ another strong baseline \textbf{m-RFT}, where the solutions which yield correct answers during sampling are used to further fine-tune the base model. 
To alleviate catastrophic forgetting, we fine-tune the model for only one epoch with a minor learning rate (1e-5). 
For comparison, we also incorporated other recent LLaMA2-base models: \textbf{MAmmoTH}~\cite{MAmmoTH} is developed by utilizing diverse datasets for math instruction, while \textbf{WizardMath}~\cite{wizardmath} employs Reinforcement Learning from Evol-Instruct (RLEIF). 
\textbf{MetaMath} 7B~\cite{metamath} is fine-tuned from the strongest English reasoning dataset, MetaMathQA.

%


%% file: latex/3.5-experiment-result.tex
\begin{table*}[t]
\setlength{\tabcolsep}{3mm}
\footnotesize
\centering
\begin{tabular}{l|cccccccccc | c}\toprule
\textbf{Model} & \textbf{Bn} & \textbf{Th}	& \textbf{Sw}	& \textbf{Ja} & \textbf{Zh}	& \textbf{Ru}	& \textbf{De}	& \textbf{Es}	& \textbf{Fr}	    & \textbf{En}   & \textbf{Avg}\\\midrule 
GPT-3.5-Turbo  & 31.2 & 38.0 & 40.0 & 36.0 & 44.0 & 43.2 & 46.0 & 47.2 & 41.6 & 54.4 & 42.2\\
\midrule
MAmmoTH 7B$^\dagger$ & 3.6 & 4.8 & 2.4 & 10.8 & 17.2 & 26.0 & 33.2 & 32.4 & 32.8 & 49.6 & 21.3 \\
WizardMath 7B$^\dagger$ & 2.0 & 4.0 & 3.4 & 24.0 & 22.4 & 30.8 & 30.4 & 34.8 & 30.4 & 47.6 & 23.0 \\
MetaMath 7B & 6.4 & 4.0 & 3.2 & 39.2 & 38.8 & 47.2 & 56.8 & 58.0 & 52.8 & 63.2 & 37.0 \\
\midrule
MathOctopus 7B & 29.2 & 33.6 & 36.4 & 35.2 & 39.2 & 38.8 & 44.8 & 42.4 & 43.2 & 52.0 & 39.5\\
+ m-RFT &25.6	&31.2&	28.8&	34.0	&39.2	&36.0	&34.8&	34.4&	36.4	&43.2	&34.4\\
+ MAPO-DPO(ours) & 30.8 & 38.0 & 37.6 & 45.2 & 47.2 & 42.0 & 45.2 & 43.2 & 40.8 & 45.6 & 41.6\\\midrule
MetaMathOctopus 7B & 25.6&	42.8	&36.4	&40.0	&46.4	&46.8	&49.6	&54.4	&46.4		&66.4 & 45.5\\
+ m-RFT & 23.2&	33.6	&34.0	&34.0	&47.2&	43.2&	45.6&	47.6	&44.8&	62.8	&41.6\\
+ MAPO-DPO(ours) & 36.0	& 44.8	& 44.8	&47.6	&55.2&	53.6&	53.6	&56.8	&52.4	& 70.8 & 51.6\\\midrule 
MistralMathOctopus 7B & 44.0 &	54.4	&50.4	&55.6 &	59.2	&58.8&	62.4	&62.0	&56.8&	76.0	&58.0 \\
+ m-RFT & 41.2	&46.8&	46.8&	48.4&	57.2	&62.8&	61.6	&59.2	&57.6&	72.0 &	55.4 \\ 
+ MAPO-DPO(ours) & \textbf{55.2}	& \textbf{60.4}&	\textbf{61.6}	&\textbf{58.0}	&\textbf{74.0}&	\textbf{70.8}&	\textbf{67.6}&	\textbf{74.0}	&\textbf{69.2}&	\textbf{82.0}	&\textbf{67.3}	\\
\midrule \midrule 

MAmmoTH 13B$^\dagger$ & 3.6 &5.2& 1.6 &19.2 &31.2& 36.8 & 45.6 & 50.0 & 39.6  &56.4& 28.9 \\
WizardMath 13B$^\dagger$ & 6.4& 5.6& 5.6& 22.0& 28.0   &  34.4&  40.4 &45.6 &42.0&  52.8& 28.3\\
MetaMath 13B & 11.6& 6.4 &7.6 &42.8 &49.2 & 63.6& 64.8 &65.2& 65.2 & 67.2 & 44.4 \\
\midrule
MathOctopus 13B &42.4&	39.2&	44.8	&38.8	&49.6	&45.2&	48.4&	53.6	&43.2		&54.8 & 46.0 \\
+ m-RFT & 29.6	&30.8	&34.4&	36.4&	40.4	&39.2&	42.0&	42.8	&40.4	&48.0&	38.4 \\ 
+ MAPO-DPO(ours)  &38.8&	46.8	&42.0&	47.6	&53.6&	49.2&	52.0&	54.4	&46.4	&54.0&	48.5	\\
\midrule
MetaMathOctopus 13B &  34.4&	42.8&	41.6&	49.2	&52.8	&54.4&	54.4	&59.2&	53.6	&71.6&	51.4\\
+ m-RFT &22.8&	29.6&	30.4	&35.2&	39.2&	40.0&	43.6&	43.6	&41.2&	59.2	&38.5	\\
+ MAPO-DPO(ours) & \textbf{44.8}	& \textbf{47.6}	&\textbf{55.2} &\textbf{56.0} &	\textbf{59.6}&	\textbf{59.2}&	\textbf{59.2}&	\textbf{63.6}&	\textbf{62.8}	&\textbf{71.6}&	\textbf{58.0}\\
\bottomrule
\end{tabular}
\caption{\label{Tab:AllLanResult-MGSM} Model Performances on MGSM test set. ``Avg'' represents the average performance in ten languages and bold text denotes the best results within the same model size. Results marked with $^\dagger$ come from \citealp{MathOctopus}.}
\end{table*}

\begin{figure*}[t]
\centering
\begin{minipage}[t]{0.45\textwidth}
\centering
\includegraphics[width=6.7cm]{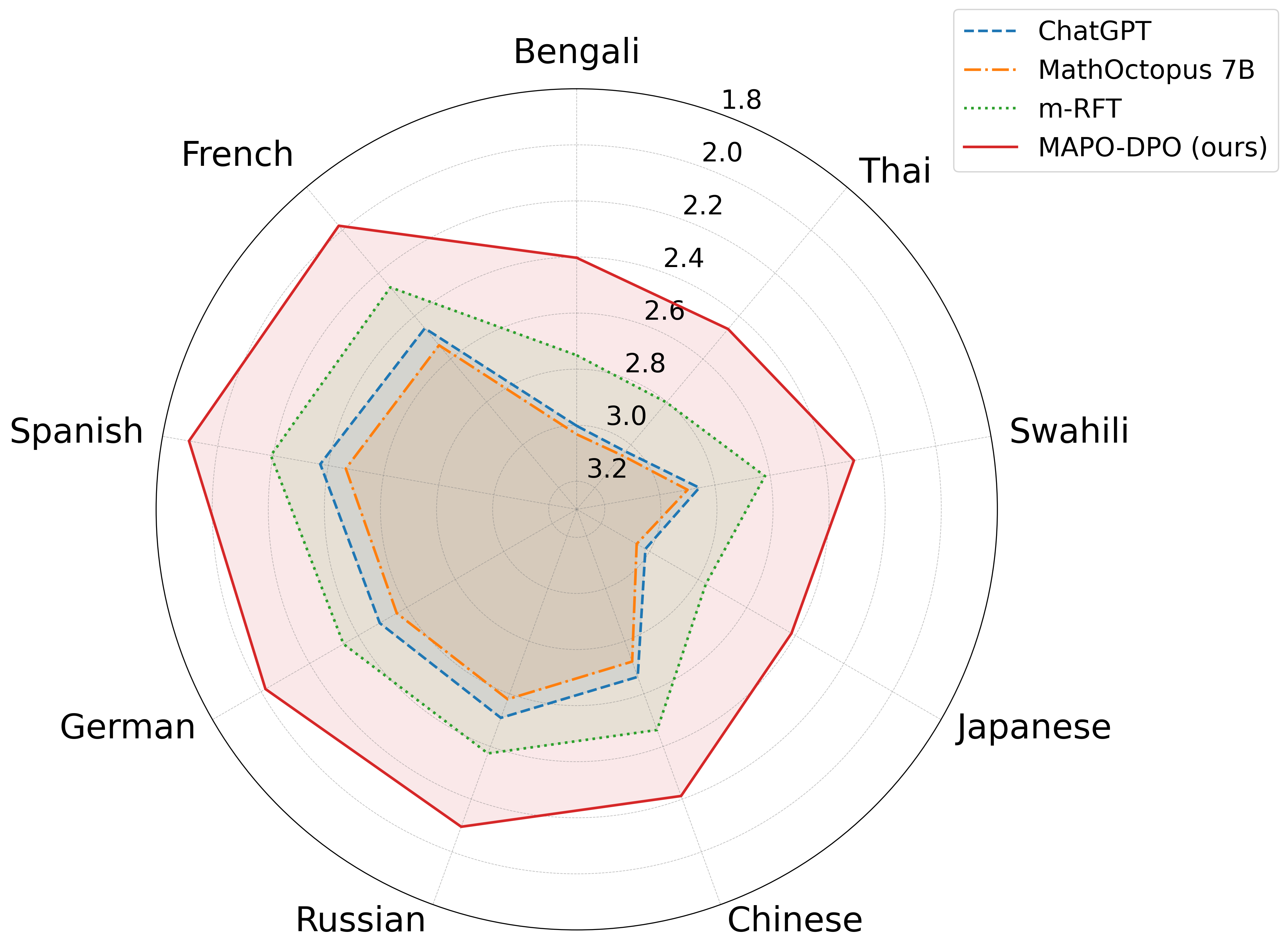}
\caption{PPL across nine languages on MSVAMP. Lower PPL indicates higher consistency in reasoning processes.}
\label{fig:ppl}
\end{minipage}
\hspace{5mm}
\begin{minipage}[t]{0.46\textwidth}
\centering
\includegraphics[width=6.7cm]{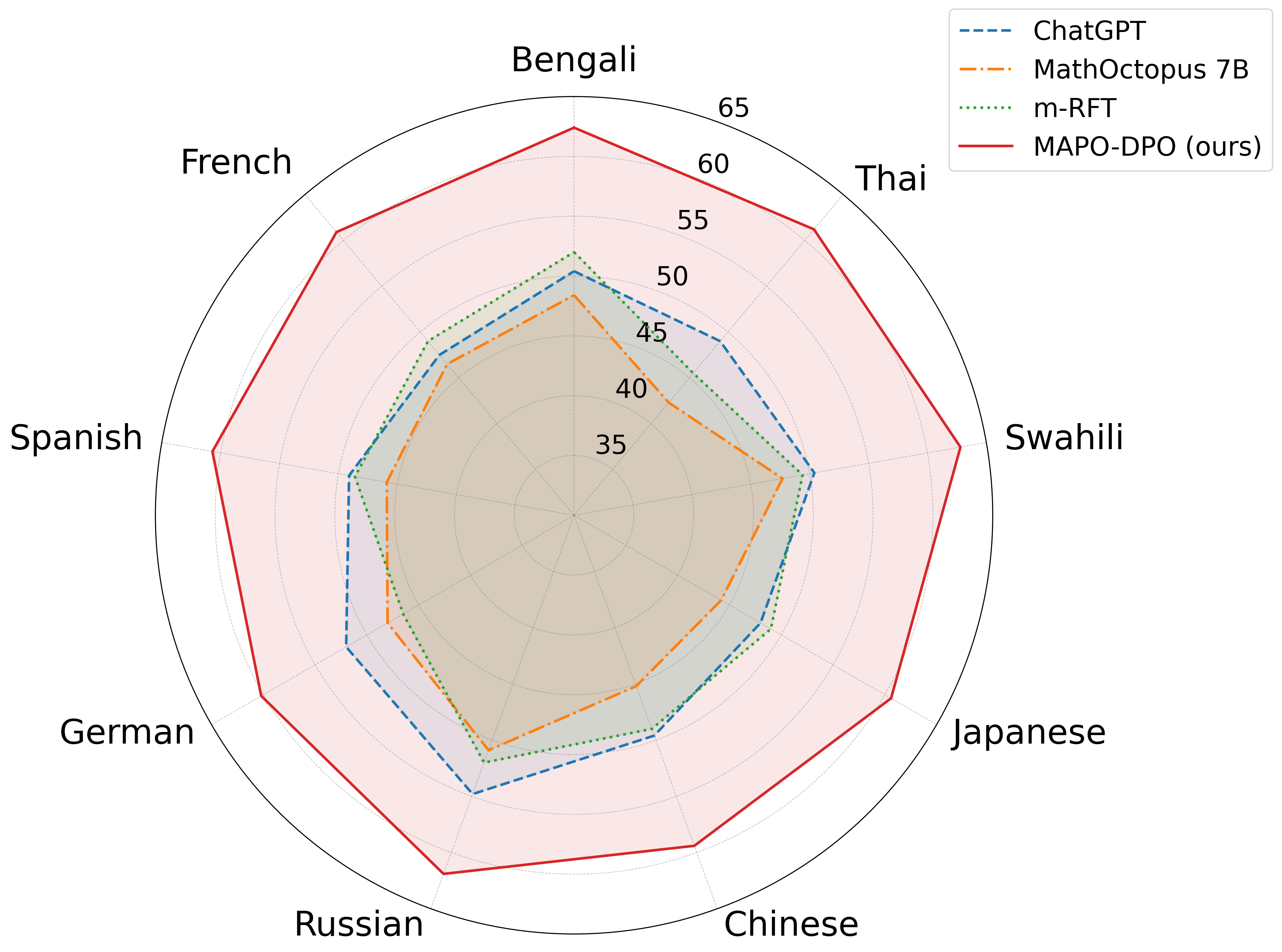}
\caption{ACR across nine languages on MSVAMP. Higher ACR indicate higher answer consistency.}
\label{fig:arc}
\end{minipage}
\end{figure*}

\subsection{Preference Optimization Improves Multilingual Reasoning Effectively}
Experimental results in Table~\ref{Tab:AllLanResult-MSVAMP},\ref{Tab:AllLanResult-MGSM},\ref{Tab:AllLanResult-NumGLUE} consistently demonstrate that our method has effectively enhanced the reasoning capabilities of the various base models and achieved state-of-the-art performance. 
More specifically, the improvement is most impressive on the out-of-domain dataset MSVAMP, where we achieved an average accuracy improvement of 16.2\% and 14.7\% MathOctopus7B and MathOctopus13B, respectively. Even for the most powerful 7B model MistralMathOctopus 7B, our method can further boost its performance to an impressive 74.6\%.
We also observe significant improvements in MGSM and MNumGLUESub. 

From the perspective of languages, most languages have improvement after alignment and it is more significant in some low-resource languages that previously under-performed. Take MathOctopus 7B as an example, our method has increased the accuracy on MSVAMP for Bengali, Thai, Swahili, and Japanese by 21.1\%, 19.3\%, 16.6\%, and 18.7\%, respectively.

Surprisingly, our preference optimization dataset does not contain English, but the accuracy of English has also improved.
After alignment, MetaMathOctopus even surpassed the English mathematical reasoning models on English questions across three datasets. 
We believe this is primarily due to the alignment facilitating a more consistent understanding of reasoning across different languages, which contributes to the enhancement of general reasoning capabilities. 

Meanwhile, our approach has also achieved stable and impressive improvements on the MetaMathOctopus 7B which has stronger reasoning capabilities, propelling it past the ChatGPT and the larger scale model MetaMathOctopus13B, demonstrating the robustness and potential of our framework.


\subsection{Alignment is the Key to Enhanced Multilingual Reasoning}



We use PPL and ACR  to evaluate the degree of alignment in reasoning processes and final answers between other languages and English, respectively.

Regarding the alignment of the reasoning process, as shown in Figure~\ref{fig:ppl}, our method effectively improves the consistency of the reasoning process, particularly for languages where the base model had poorer alignment, such as Bengali, Thai, and Japanese.
This proves that the model can show more similar reasoning thinking to English on non-English after alignment.



Additionally, we notice that alignment also contributes to more consistent final answers.
A higher ACR indicates that there is a greater overlap between the questions answered correctly in non-English and those answered correctly in English.
From the results in Figure~\ref{fig:arc}, our alignment framework has greatly increased the ACR for each language. 
This means that the performance gains in Table~\ref{Tab:AllLanResult-MSVAMP} stem from the parts that intersect with English. 
These observations demonstrate that our method effectively aligns the model's non-English reasoning processes with English, thereby enhancing reasoning abilities in non-English languages.

\subsection{Generalizable Multilingual Reasoning}

As indicated in Table~\ref{Tab:AllLanResult-MSVAMP}, on the out-of-domain test set, MAPO achieved a 16.2\% and 11.7\% increase in accuracy, which is significantly higher than the 7.5\% and 3.7\% achieved by m-RFT. 
Surprisingly, while MAPO effectively improves the multilingual reasoning performance on both two datasets, it also improves the performance on MGSM. 
In contrast, m-RFT exhibited a 5.1\% and 3.9\% performance degradation.

We believe that this is mainly because the model directly learns the given labels during SFT. 
The labels inevitably involve data-specific attributes, which are difficult to generalize. 
Conversely, our method does not give the model ``correct answers'', but instead guides the model to generate outputs we prefer, which is more effective in allowing the model to learn multilingual reasoning, and this preference can better generalize across various challenges.

%% file: latex/4-analysis.tex
\begin{table}[t]
\setlength{\tabcolsep}{2mm}
\footnotesize
\centering
\begin{tabular}{lcc}\toprule
 &  \textbf{MGSM}  & \textbf{MNumGLUESub}\\\midrule
\text{Acc. on MSVAMP} & 31.6 & 54.9\\
\text{Self-BLEU(En)} & 0.81 & 0.53 \\
\text{Self-BLEU(Non-En)} & 0.84 & 0.60 \\
\bottomrule
\end{tabular}
\caption{\label{DataInfo} Comparison between  MGSM8KInstruct (MGSM) and MNumGLUESub. We report the average Self-BLEU scores for the sampled reasoning in English and non-English. 
}
\end{table}

\subsection{Preference Estimating on a Dataset Different from SFT}
We conduct experiment with preference optimization using the MGSM8KInstruct dataset instead of the MNumGLUESub. Table~\ref{DataInfo}  present the average accuracy over 10 languages on MSVAMP. 
The results indicate that preference optimization based on the SFT dataset lead to the performance degradation on the out-of-domain test set MSVAMP. We believe this is because SFT make the model quickly remember the oracle reasoning process, which leads to a serious degradation in the generation diversity. Thus, the policy model struggles to learn the alignment preferences we truly hope to optimize and easily falls into the trap of overfitting.

For further analysis, We randomly sample 1000 reasoning processes of LLMs on the MGSM8KInstruct and MNumGLUESub, and calculated the Self-BLEU~\cite{selfbleu} on sampled reasoning process of the same problem. The results (Table~\ref{DataInfo}) suggest that on MGSM dataset, LLMs tends to generate quite similar reasoning, compared with MNumGLUESub, which confirms the diversity degradation.

\begin{table}[t]
\setlength{\tabcolsep}{2mm}
\footnotesize
\centering
\begin{tabular}{lccc}
\toprule
System                & MSVAMP       & MGSM  \\
\midrule
MetaMathOctopus7B    & 53.0 & 45.5	  \\
+  Ours \& NLLB 600M & 61.1	 & 48.9	  \\
+  Ours \& MBART-MMT-600M          & 59.9 &	49.6	   \\
+  Ours \& M2M-1.2B        & 61.4 &	49.3	   \\
\bottomrule
\end{tabular}
\caption{Average accuracy on ten languages on MSVAMP and MGSM. Note that we only conduct one iteration MAPO-DPO with different translation models.\label{tab:robustnessDiffMT}}
\end{table}

\begin{table*}[t]
\setlength{\tabcolsep}{3mm}
\footnotesize
\centering
\begin{tabular}{c|p{12cm}}\toprule
\textbf{English Question} & 182 students suggested adding mashed potatoes while others suggested adding bacon to the menu . if 166 more students suggested adding bacon than those that suggested mashed potatoes how many students suggested bacon ?\\\midrule 
\textbf{Chinese Question} & 
\kaiti{182名学生建议在菜单中加入土豆泥，而其他人建议加入培根。如果有比建议加入土豆泥的学生多166人建议加入培根，那么有多少学生建议加入培根？} \\
\midrule 
\textbf{English Solution}  & The number of students who suggested mashed potatoes is 182. The number of students who suggested bacon is 182+166=348. \textcolor{green}{\ding{51}} \\
 \midrule
 \textbf{Chinese Solution}  & \textbf{[Before Alignment]} \\
 &\kaiti{建议加入土豆泥的学生比建议加入培根的学生多166人，所以两边减去166得到差值为182 - 166 = 16，因此，建议加入培根的学生人数为16。\color{gray}(There are 166 more students recommending mashed potatoes than bacon so subtract 166 from both sides to get a difference of 182-166 = 16. Therefore, the number of students recommended to join Bacon is 16.)} \textcolor{red}{\ding{55}} \\
 &\textbf{\kaiti{[PPL]:}}  2.65  \\
 &  \\
  &  \textbf{[After Alignment]} \\
 & \kaiti{建议加入土豆泥的学生人数是182。建议加入培根的学生人数是182 + 166 = 348。} \color{gray}(The number of students who suggested mashed potatoes is 182. The number of students who suggested bacon is 182+166=348.) \textcolor{green}{\ding{51}} \\
 &\textbf{\kaiti{[PPL]:}}  0.97  \\
\bottomrule
\end{tabular}
\caption{\label{example} An example from the MSVAMP test set. The example clearly demonstrates that, through our preference optimization method, the model successfully corrects erroneous reasoning processes and generates reasoning processes that is more consistent with English.
}
\end{table*}

\section{Robustness over Different Translation Models}
To verify the robustness of our framework across various translation models, we conduct experiments using MBART-MMT-600M~\footnote{\url{https://huggingface.co/facebook/mbart-large-50-many-to-many-mmt}} and M2M-1.2B~\footnote{\url{https://huggingface.co/facebook/m2m100\_1.2B}}, which have the different model architectures, scales, and training datasets. The results, presented in Table~\ref{tab:robustnessDiffMT}, demonstrate that our framework consistently achieves performance improvements. The performance is even slightly enhanced when with M2M-1.2B. These experimental results conclusively confirm the robustness of our framework.  In addition, we have conducted experiments on translation models of different scales, and the results also demonstrate the robustness of our method. For more details, please refer to the Appendix~\ref{ScalingMT}.

\begin{figure}[t]
\centering
\includegraphics[scale=0.25]{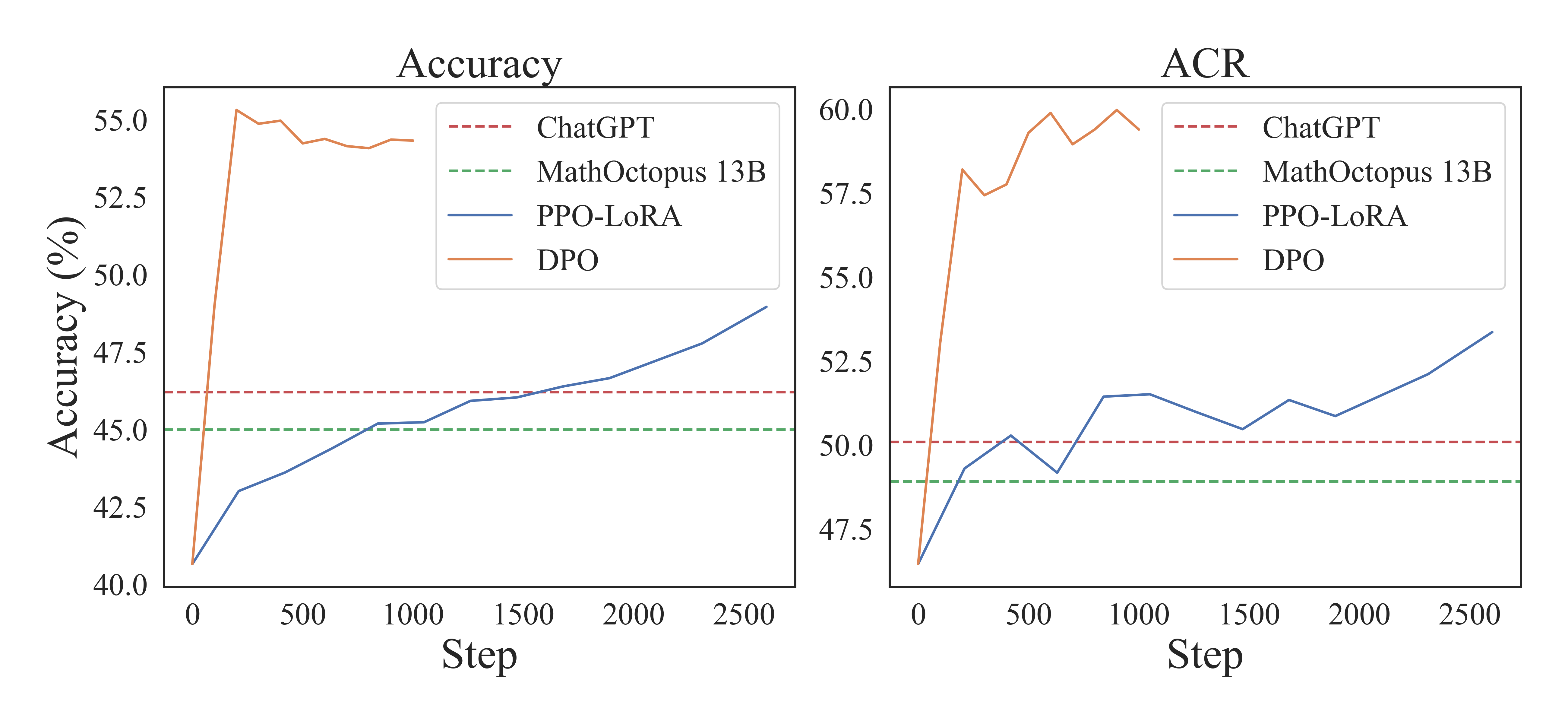}
\caption{Average accuracy on Non-English math problems and ACR versus preference optimization step on MSVAMP. MathOctopus 13B and ChatGPT are selected for comparison. }
\label{trainDynamic}
\end{figure}

\subsection{Reasoning alignment during Preference Optimization}

To better analyze the optimization preference process, we tested the checkpoints of the DPO and PPO-LoRA on MSVAMP and visualized the accuracy and ACR respectively in Figure~\ref{trainDynamic}.

The results show that the optimization of DPO is quite efficient. DPO enabled the MathOctopus 7B model to achieve higher accuracy and ACR than both ChatGPT and MathOctopus 13B within just 100 steps. As the optimization continues, the accuracy remains relatively stable while the degree of alignment still shows an increasing trend. 

Meanwhile, although the optimization speed of PPO is slower, it take fewer than 1000 steps to surpass MathOctopus 13B. 
As the model continued to explore and optimize the preferences, it achieved superior multilingual reasoning performance compared to ChatGPT at 1600 steps.

The experimental results demonstrate that the model continuously aligns non-English reasoning with English reasoning, effectively enhancing its multilingual reasoning capabilities. 
For PPO-LoRA, although optimization is halted around 2500 steps due to the limitations on computational resources, we notice that growth in multilingual reasoning consistency and ability continues even beyond 2500 steps, indicating the potential of our preference optimization framework.

\subsection{Alignment Reasoning for Improvements}

The example in Table~\ref{example} directly illustrates how our framework can improve reasoning ability by aligning reasoning in other languages to English.
For the given English problem, the basic model correctly analyzes that an addition should be conducted to find the number of students recommended to join Bacon. 
However, when given the same problem in Chinese, LLM misjudges the relationship of variables and writes a significantly different reasoning process. 
This directly proves that although the baseline model has been fine-tuned with multilingual reasoning data, its reasoning and thoughts are inconsistent, whereas non-English reasoning is more prone to errors. 
After alignment, the reasoning thoughts are more similar to the English answer, and the reasoning thoughts are also corrected.
